\documentclass[10pt,twocolumn,letterpaper]{article}

\usepackage{wacv}
\usepackage{times}
\usepackage{epsfig}
\usepackage{graphicx}
\usepackage{amsmath}
\usepackage{amssymb}

\usepackage[normalem]{ulem}

\wacvfinalcopy 

\ifwacvfinal\fi
\begin{document}

\title{Accurate 6D Object Pose Estimation by Pose Conditioned Mesh Reconstruction}

\author{Pedro Castro \hspace{2cm} Anil Armagan \hspace{2cm} Tae-Kyun Kim \\
Imperial College London\\
{\tt\small {p.castro18, a.armagan, tk.kim}@imperial.ac.uk}
}
\def\eg{\emph{e.g.}}
\def\Eg{\emph{E.g.}}
\def\etal{\emph{et. al. }}
\def\wrt{w. r. t. }

\newcommand{\image}{\mathcal{I}}
\newcommand{\imagecrop}{\mathcal{I}_{ROI}}
\newcommand{\maskgt}{\mathcal{H}_{m}}
\newcommand{\mask}{\hat{\mathcal{H}}_{m}}
\newcommand{\centroidgt}{\mathcal{H}_{c}}
\newcommand{\centroid}{\hat{\mathcal{H}}_{c}}
\newcommand{\mesh}{\mathcal{\hat{M}}}
\newcommand{\meshcon}{\mathcal{M}}
\newcommand{\rotationallo}{\mathcal{R}_{a}}
\newcommand{\rotation}{\mathcal{R}}

\newcommand{\pose}{\mathcal{P}_{6D}}

\maketitle

\begin{abstract}
Current 6D object pose methods consist of deep CNN models fully optimized for a single object but with its architecture standardized among objects with different shapes. In contrast to previous works, we explicitly exploit each object's distinct topological information i.e. 3D dense meshes in the pose estimation model, with an automated process and prior to any post-processing refinement stage. In order to achieve this, we propose a learning framework in which a Graph Convolutional Neural Network reconstructs a pose conditioned 3D mesh of the object. A robust estimation of the allocentric orientation is recovered by computing, in a differentiable manner, the Procrustes' alignment between the canonical and reconstructed dense 3D  meshes. 6D egocentric pose is then lifted using additional mask and 2D centroid projection estimations. Our method is capable of self validating its pose estimation by measuring the quality of the reconstructed mesh, which is invaluable in real life applications. In our experiments on the LINEMOD, OCCLUSION and YCB-Video benchmarks, the proposed method outperforms state-of-the-arts.
\end{abstract}

\section{Introduction}

Estimating accurate 6D pose of an object has been an important problem to deal with as it raised many applications such as augmented reality or robotics. Several problems exist as 6D pose space is huge and objects might appear symmetric, and under challenging conditions such as occlusion and illumination.
Many approaches have been proposed to attack the problems. Some approaches rely on depth sensors and exploit the depth information which is essential for 6D pose estimation \cite{nocs, depth_only, linemod, densefusion, aly}. Other approaches use monocular RGB images to estimate the 6D pose of an object where the problem can be more challenging \cite{segmentationdriven, bb8, seamless, posecnn, heatmaps}.

Previously, learning based techniques for 6D object pose estimation perform feature learning \cite{aae}, direct pose regression \cite{posecnn, 3drcnn}, or 3D bounding box corner regression \cite{bb8, seamless, nocs}, for instance. They required the models to indirectly learn the rigid object's geometry even though its 3D CAD model is known. 
More recently, Wang \textit{et. al.} \cite{nocs} introduce a new object surface coordinate representation and propose a dense per-pixel estimation of surface points in 2D space. This change in paradigm i.e. to dense pose estimation is analogous to the one seen in human body \cite{densebody}, face \cite{denseface, facedense2} and hand \cite{densehand} pose estimation. Suwajanakorn \textit{et. al.} \cite{keypoints} learned the best 3D surface keypoints for pose estimation, pursuing keypoints geometrically and semantically consistent from different viewpoint. Whereas target objects in other problems are in a categorical level \cite{shapelearning, keypoints, 3drcnn, densebody}, deformable and/or articulated, objects in 6D pose estimation are typically in an instance level and rigid. 

\newcommand*{\vcenterimage}[1]{\vcenter{\hbox{\includegraphics[width=0.27\linewidth]{#1}}}}
\newcommand*{\vcenterarrow}{\vcenter{\hbox{$ \Longrightarrow $}}}


\begin{figure}[t]
\centering
$\vcenterimage{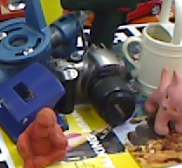}\hspace{3pt}\vcenterarrow\hspace{3pt}\vcenterimage{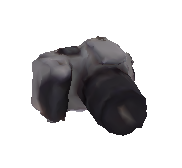}\hspace{3pt}\vcenterarrow\hspace{3pt}\vcenterimage{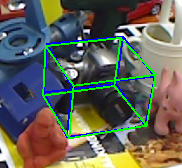}$ \\
\vspace{5pt}
(a) \hspace{0.31\linewidth} (b)\hspace{0.31\linewidth} (c)  \\
\vspace{5pt}
\hrule 
\vspace{2pt}
$\vcenterimage{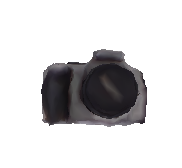}\hspace{0.09\linewidth}\vcenterimage{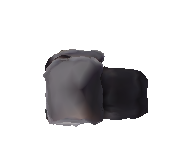}\hspace{0.045\linewidth}\vline\hspace{0.045\linewidth}\vcenterimage{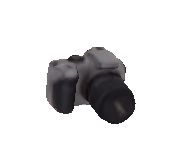}$  \\
\vspace{2pt}
(d) \hspace{0.31\linewidth} (e)\hspace{0.31\linewidth} (f)  \\
\vspace{2pt}
\vspace{5pt}
\caption{Our method recovers the 6D pose of an object from a single image by means of an intermediate object reconstruction. From an RGB image of a detected object (a), a pose conditioned mesh (b) is reconstructed from which 6D pose can be extracted (c). (d) and (e) present alternative viewpoints of reconstruction (b) while (f) displays the groundtruth mesh. }
\vspace{-10pt}
\end{figure}

In this paper, we propose a deep architecture that recovers an object's 6D pose from the input of a single 2D RGB image via 3D reconstruction. The object's 3D topological information is directly introduced in our model's architecture, disentangling the learning of the object's pose from its shape. 
In turn, this allows us to explicitly exploit object 3D dense shapes for pose estimation, different from prior arts, learning sparse 3D keypoints or estimating dense per-pixel poses. Unlike previous methods, which fit an object 3D keypoints to correspondent 2D predictions in a {\it post processing} step \cite{seamless, bb8, nocs}, we make to use the object's 3D model information prior to prediction. In our problem, object 3D shapes are rigid and known \textit{a priori}, thus full 3D shape supervisions (cf. 2D mask supervisions \cite{sb, 3drcnn, handgraphcnn}) are available, and learning/inferring pose-dependent shapes is more affordable.

\begin{figure*}[t] 
\begin{center}
\includegraphics[width=\linewidth]{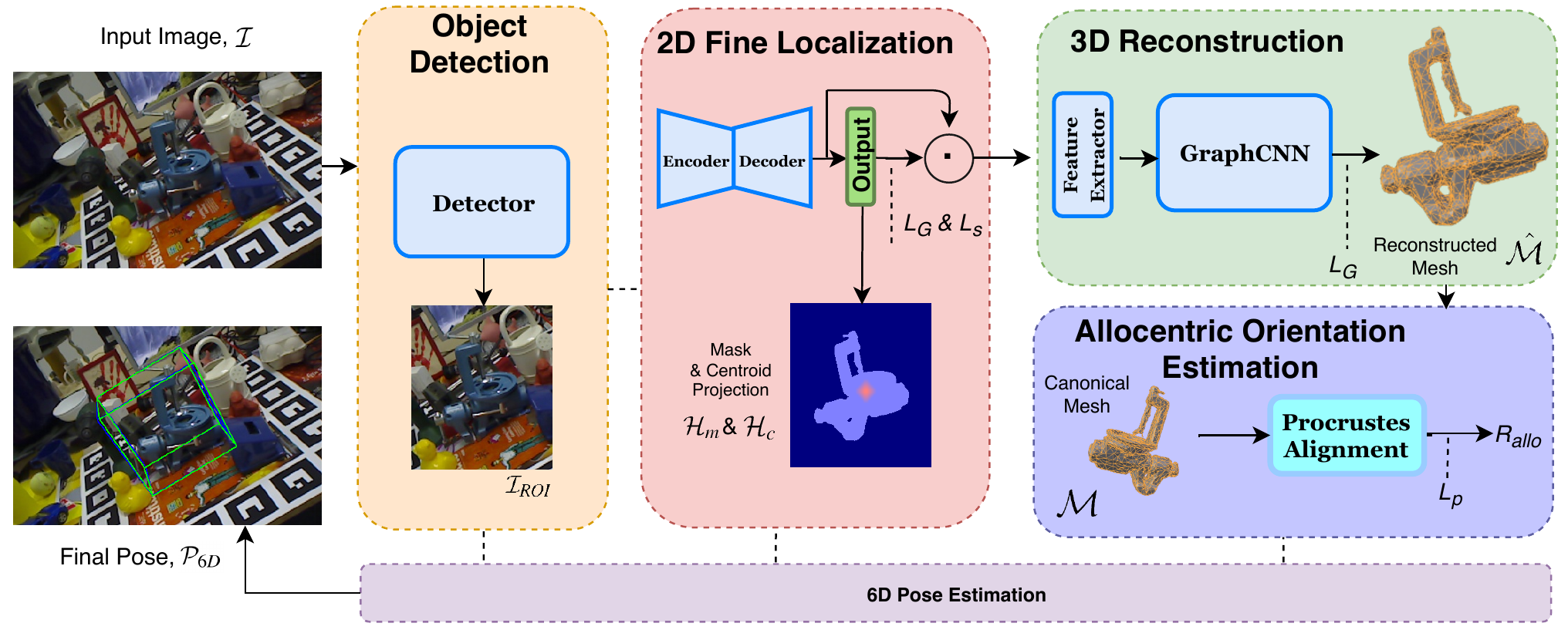}
\end{center}
   \caption{Overview of our pipeline where we fully exploit the object shape topology both in 2D and 3D for 6D pose estimation. Given an RGB input image $\image$, an object ROI, $\imagecrop$, is cropped with a pretrained 2D object detector (orange). The cropped region is used to finely localize the object by estimating a binary object mask, $\mask$ and a heatmap, $\centroid$, for the object centroid (pink) and use them later to reconstruct a pose conditioned 3D mesh $\mesh$ of the object (green). An allocentric orientation $\rotationallo$ of $\mesh$ \wrt a corresponding canonical object mesh $\meshcon$ can be estimated (purple) in a differentiable way by finding the relative rotation of $\mesh$ to a corresponding canonical object mesh, $\meshcon$, with the Procrustes's alignment. Cropped image locations, $\mask$, $\centroid$ and $\rotationallo$ are later used to estimate the full 6D pose of the object. $\bigodot$ represents the channel-wise concatenation operation. }
\label{fig:pipeline}
\end{figure*}

The proposed method, shown in Figure \ref{fig:pipeline}, is primarily composed of three main components, \textbf{1)} a 2D fine localization module which produces both a segmentation mask and the 2D projection of the object's centroid, following an off-the-shelf detector, \textbf{2)} a GraphCNN \cite{graphcnn} which learns to fit a pose conditioned dense 3D mesh of the object of interest, and lastly \textbf{3)} the estimation of object allocentric orientation by computing the optimal rotation necessary to align the produced mesh to the object's canonical mesh. These three main components are jointly learnt in an end-to-end manner. Given a previously estimated padded 2D bounding box, centroid projection, mask segmentation and allocentric pose, along with the camera's intrinsic parameters, we lift the 6D egocentric pose of the object w.r.t the camera.

Our main contributions are: 
\begin{itemize}
    \item We introduce a novel reconstruction based approach for 6D object pose estimation, which takes advantage of each object topological information. We learn a distinct per-object GraphCNN to recover an accurate pose conditioned reconstructed mesh of the target object.
    \item We demonstrate how our method is able to outperform previous works in both the standard 6D benchmark dataset LINEMOD as well as on more challenging datasets such as the OCCLUSION and YCB-Video datasets.
    \item We developed a self-validating technique allowing our method to evaluate its pose estimation accuracy at test time.
\end{itemize}

\section{Related Work}

In this section we present a review of previous work on 6D object pose estimation as well as from other pose estimation fields relevant to our method.

\textbf{Classical.} Classical 6D object pose relied on template-matching \cite{linemod}, local descriptors \cite{local_descriptors} and 3D object alignment with Chamfer edge matching \cite{chamfer} or Hausdorff distance \cite{hausdorff}. Although very fast and somewhat robust to occlusion, these methods often struggle on poorly textured objects and low resolution images. Some recent non-learning based methods were shown effective in the 6D object pose challenge \cite{bop}.

\textbf{RGB-D.} Nowadays, depth information is more easily available with the dispersal of cheap depth sensors and adoption of multi-camera setups. Naturally, a line of methods have focused on 6D pose estimation from RGB-D \cite{linemod,keypoints1, depth1, densefusion, aly, andreas} or depth information only \cite{depth_only}. Additionally, applying a refinement step in the form of standard ICP \cite{bb8, ssd6d} is a common practice, though its success depends heavily on correct initialisations, and tuning its hyper-parameters is not always straightforward. In this work, we focus on recovering full 6D pose from a single RGB image. 

\textbf{Deep Learning.} Over the last years, deep learning based techniques have been dominating the 6D research field. Detection is a crucial part of recovering an object's 6D pose and the field of 2D detection has been the focus of remarkable works \cite{ssd, yolo, fasterrcnn}. Therefore, relying on top of the quality of existing 2D detection techniques, a plethora of works propose a cascade of modules \cite{bb8, aae, nocs}, where an initial detection of the target object is left to pre-existing competent detectors with the consequent modules focusing on additional pose tasks. 
The most straight forward way of recovering an object's pose is by estimating its pose parameters directly \cite{posenet, posecnn, 3drcnn}. In particular, \cite{posecnn} decouples translation and orientation estimation and \cite{3drcnn} adds an additional shape recovering task in their end-to-end pose estimator. Going further, \cite{ssd6d} disentangles the target's allocentric orientation from the camera's inplane rotation. Similarly to us, they compare the pre-processed bounding box sizes with estimated ones to recover the object's distance to the camera. However, in order to yield acceptable results, they rely on a complex RGB refinement step. In general, direct pose estimation faces problems when dealing with particular unseen poses and occlusion \cite{tripletpose}. Using an autoencoder to its full potential, \cite{aae} creates a database of pose conditioned encoded latent vectors. At test time, the image is encoded into the latent space and the pre-computed pose database is searched for the closest match.

\textbf{Keypoint-Based Methods.} In more recent works, direct pose estimation is replaced by predicting the 2D pixel coordinate of the 3D bounding box's corners \cite{bb8, seamless, segmentationdriven, heatmaps}. From these control points, pose can be efficiently extracted using openly available implementations of the PnP algorithm \cite{epnp}. While \cite{bb8} proposes a three stage pipeline where each stage is independently optimized, including refinement,
\cite{seamless} extends the power of the YOLO \cite{yolo} architecture and predicts on each cell the keypoint locations, in addition to the existing classification and bounding box outputs. 
Occlusion is a serious challenge for pose estimators since it introduces foreground distractions which may hide critical features. Works like \cite{heatmaps, segmentationdriven} have focused on alleviating occlusion by locating 2D keypoints over multiple patches over the image.
We are starting to see a resurging of an old technique where we replace sparse keypoint prediction with dense per-pixel. The recent work by Wang \textit{et. al.}{} \cite{nocs} introduces an new object surface coordinate representation and proposes a dense per-pixel estimation of surface points in 2D space. This change in paradigm is analogous to the one seen in the fields of human body \cite{densebody} and hand \cite{densehand} pose estimation. 

While these hand-crafted keypoints and surface representation have been proven to work, \cite{keypoints} are motivated to learn the optimal surface keypoints for pose estimation without any keypoint annotation supervision. The set of learnt optimal 3D keypoints are geometrically and semantically consistent from any viewpoint. They learn to predict the same set of 3D keypoints from 2 different views, recover the relative pose between views by aligning the 2 sets of 3D keypoints \cite{procrustes} and optimize their model by minimize the relative pose error. Like them, we are interested in recovering the pose by aligning two set of points, and in our case, these are the estimated 3D points of a rigid mesh with the canonical position of the corresponding points.

\textbf{Other 3D shape domains.} In other fields, some methods have taken advantage of the graph-like structure of 3D meshes. In the face reconstruction domain, \cite{facegraphcnn} proposed a mesh based autoencoder, whose operations are applied to the face encoded as a mesh surface. \cite{handgraphcnn} proposed a similar approach in the hand pose domain, reconstructing a full 3D hand mesh using spectral convolutions \cite{graphcnn}. Since an object's mesh can also be represented by a graph and inspired by the success of these approaches, we purpose a GraphCNN capable of estimating a pose conditioned 3D mesh, taking advantage of the network's ability to exploit known topological information.

\section{Methodology}

In instance-level 6D object pose scenarios, full object 3D information is usually provided prior to pose inference \cite{heatmaps, aae}. Recent trends show that learning to detect object keypoints based on their appearance can achieve state of the art results \cite{seamless, bb8}. However, even these techniques indirectly learn the shape of the object, shown by their ability to accurately predict occluded keypoints.Therefore, understanding the shape of an object is crucial for any pose estimation technique. 

Given a monocular RGB input image $\image$, our goal is to estimate full 6D pose of a rigid object. We aim to design a distinct per object architecture in an automated manner by taking full advantage of prior information of the object. Inspired by previous approaches~\cite{handgraphcnn,pix2mesh}, our reconstruction stage combines the use of the object's known topology with encoded pose information extracted from $\image$. Afterwards, estimated mesh information is used to recover the allocentric orientation \cite{keypoints} of the target object. Egocentric orientation can then be recovered and lifted to 6D by adopting different approaches from the literature~\cite{3drcnn, ssd6d}.

We use a pretrained FasterRCNN~\cite{fasterrcnn} based 2D object detector and fine tune the model on our training data in order to detect an object in 2D space. The detector is used to crop an object region-of-interest (ROI) $\imagecrop$ for further processing. $\imagecrop$ is used in a high resolution to extract fine details of object appearance in the next stages of our pipeline. This \textit{ad hoc} detector is trained independently.


Our main motivation is similar to discovery of latent keypoints for object pose estimation~\cite{keypoints}, backed by the success of dense mesh estimators~\cite{densebody, handgraphcnn}. We infer the 3D keypoints position by reconstructing a 3D object mesh, $\mesh$. Section~\ref{reconstruction} discusses how we learn to estimate the mesh vertices with a GraphCNN~\cite{graphcnn}. 

Exploited cues, $\mask$, $\centroid$ and $\mesh$ of the object are crucial for our pipeline to make accurate 6D pose estimation. We use these cues to estimate an allocentric orientation of $\mesh$ and later to lift this representation into a 6D egocentric pose \wrt the camera. Our allocentric orientation learning is inspired from~\cite{keypoints} but we combine it to recover the orientation of $\mesh$ by aligning its keypoints with a canonical mesh, $\meshcon$, in a differentiable manner, as discussed in Section~\ref{lifting}.

In the rest of this section, the 3 main stages of our 6D object pose estimation pipeline are discussed in greater detail.



\subsection{Fine Object Localization} \label{segnet}
As an important piece of our pipeline we want to make fine localization of the object's appearance. Localization cues exploited in this section caries essential information for pose estimation and integral to lift 6D pose which will be discussed later in Section~\ref{lifting}.

Given a $\imagecrop$, 
we estimate both a binary segmentation mask $\mask$ of object's silhouette and a heatmap $\centroid$ for the 2D projection of the object's centroid. Due to the precise nature of the 6D pose estimation task, both $\centroid$ and $\mask$ are estimated at a high resolution making our method robust to the detector's inaccuracies.
We learn to estimate 2D projection of the object's centroid as a heatmap $\centroid$ around the projection. $\centroid$ is defined as a 2D Gaussian distribution centered on the expected 2D projection of the centroid with $\sigma=5px$, this way preserving translation equivariance w.r.t. the image. Please note that the object centroid is defined as center of the object in the model space and the 2D projection can be easily obtained given the camera intrinsics. 

Inspired by \cite{multiscale}, we learn to estimate $\mask$ and $\centroid$ by minimizing a multi-loss function $L_{s}$ which is constructed from multiple scales and stages of the decoders. In practice, we make use of three scales. At test time, outputs from the scale with the highest resolution are used for pose estimation. We use $\mask^{(i)}$ and $\centroid^{(i)}$ as our estimations during inference for $\mask$ and $\centroid$ where $i$ corresponds to the index for the highest scale among a set of scales $S$. More formally, we minimize the loss function $L_{s}$:

\begin{equation}
\label{lscale}
    L_s = \sum_{i=0}^{\mid S \mid} \frac{\lambda_m L_m^{(i)} + \lambda_c L_c^{(i)}}{2^i}
\end{equation}
where $i$ is the index of outputs at different scales $S$ and $\lambda_m$ and $\lambda_c$ are the weight ratios for the mask and the centroid losses.

$L_{m}$ and $L_{c}$, the mask and 2D projection losses at each scale $S_i$, are defined as a binary cross entropy and mean squared error functions, respectively:
\begin{multline}
    L_{m}^{(i)} =  \frac{1}{\lvert\maskgt^{(i)}\rvert}\sum_{p} - (\maskgt^{(i)}(p)  \cdot log(\mask^{(i)}(p)) + \\ (1 - \maskgt^{(i)}(p)) \cdot log(1 - \mask^{(i)}(p)))
\end{multline}

\begin{equation}
    L_{c}^{(i)} = \sum_{p} {\lVert { \centroid^{(i)}(p) - \centroidgt^{(i)}(p) } \rVert}^2_2
\end{equation}
where $\maskgt^{(i)}$ and $\centroidgt^{(i)}$, the groundtruth mask and 2D centroid heatmap with their corresponding predictions $\mask^{(i)}$ and $\centroid^{(i)}$ are used for each pixel location $p$ within a length of $\lvert\mask^{(i)}\rvert$.

\newcommand\imageWidth{1.8cm}
\newcommand\imageHeight{1.8cm}
\begin{figure*}[!h]
\begin{tabular}{c|c|c}

\begin{tabular}{ccc}
\hspace{-10pt} 
\includegraphics[width=\imageWidth,height=\imageHeight]{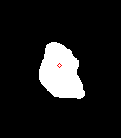} & \hspace{-15pt} 
\includegraphics[width=\imageWidth,height=\imageHeight]{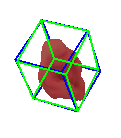} \hspace{-15pt} &
\includegraphics[width=\imageWidth,height=\imageHeight]{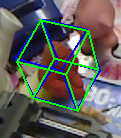} \hspace{-10pt} \\
\hline\hspace{-10pt} 
\end{tabular}

&
\begin{tabular}{ccc}
\hspace{-10pt} 
\includegraphics[width=\imageWidth,height=\imageHeight]{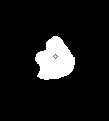} & \hspace{-15pt} 
\includegraphics[width=\imageWidth,height=\imageHeight]{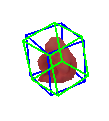} \hspace{-15pt} &
\includegraphics[width=\imageWidth,height=\imageHeight]{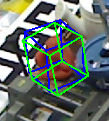} \hspace{-10pt} \\
\hline\hspace{-10pt} 
\end{tabular}

&
\begin{tabular}{ccc}
\hspace{-10pt} 
\includegraphics[width=\imageWidth,height=\imageHeight]{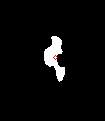} & \hspace{-15pt} 
\includegraphics[width=\imageWidth,height=\imageHeight]{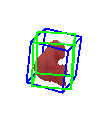} \hspace{-15pt} &
\includegraphics[width=\imageWidth,height=\imageHeight]{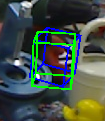} \hspace{-10pt} \\
\hline\hspace{-10pt} 
\end{tabular} \vspace{-7pt} \\ 

\begin{tabular}{ccc}
\hspace{-10pt} 
\includegraphics[width=\imageWidth,height=\imageHeight]{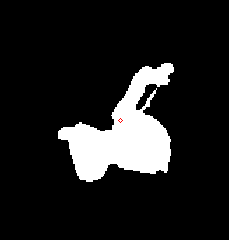} & \hspace{-15pt} 
\includegraphics[width=\imageWidth,height=\imageHeight]{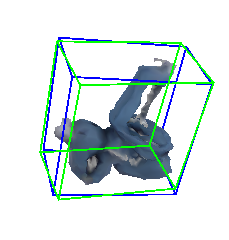} \hspace{-15pt} &
\includegraphics[width=\imageWidth,height=\imageHeight]{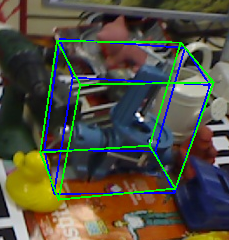} \hspace{-10pt} \\
\hline\hspace{-10pt} 
\end{tabular}

&
\begin{tabular}{ccc}
\hspace{-10pt} 
\includegraphics[width=\imageWidth,height=\imageHeight]{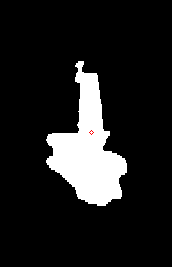} & \hspace{-15pt} 
\includegraphics[width=\imageWidth,height=\imageHeight]{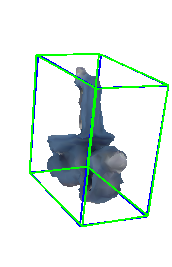} \hspace{-15pt} &
\includegraphics[width=\imageWidth,height=\imageHeight]{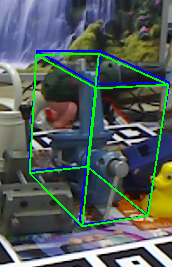} \hspace{-10pt} \\
\hline\hspace{-10pt} 
\end{tabular}

&
\begin{tabular}{ccc}
\hspace{-10pt} 
\includegraphics[width=\imageWidth,height=\imageHeight]{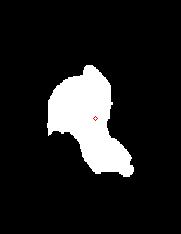} & \hspace{-15pt} 
\includegraphics[width=\imageWidth,height=\imageHeight]{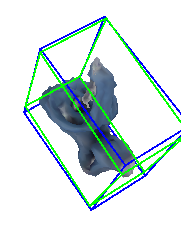} \hspace{-15pt} &
\includegraphics[width=\imageWidth,height=\imageHeight]{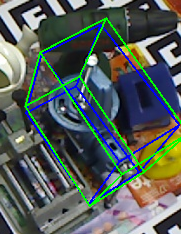} \hspace{-10pt} \\
\hline\hspace{-10pt} 
\end{tabular} \vspace{-7pt} \\

\begin{tabular}{ccc}
\hspace{-10pt} 
\includegraphics[width=\imageWidth,height=\imageHeight]{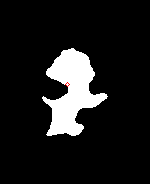} & \hspace{-15pt} 
\includegraphics[width=\imageWidth,height=\imageHeight]{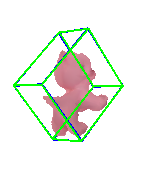} \hspace{-15pt} &
\includegraphics[width=\imageWidth,height=\imageHeight]{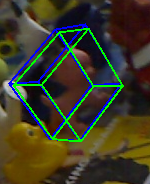} \hspace{-10pt} \\
\hline\hspace{-10pt} 
\end{tabular}

&
\begin{tabular}{ccc}
\hspace{-10pt} 
\includegraphics[width=\imageWidth,height=\imageHeight]{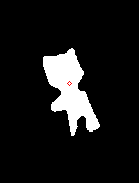} & \hspace{-15pt} 
\includegraphics[width=\imageWidth,height=\imageHeight]{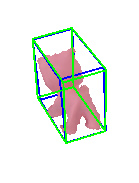} \hspace{-15pt} &
\includegraphics[width=\imageWidth,height=\imageHeight]{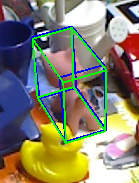} \hspace{-10pt} \\
\hline\hspace{-10pt} 
\end{tabular}

&
\begin{tabular}{ccc}
\hspace{-10pt} 
\includegraphics[width=\imageWidth,height=\imageHeight]{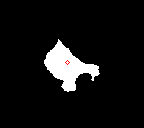} & \hspace{-15pt} 
\includegraphics[width=\imageWidth,height=\imageHeight]{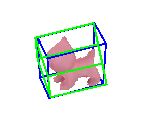} \hspace{-15pt} &
\includegraphics[width=\imageWidth,height=\imageHeight]{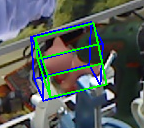} \hspace{-10pt} \\
\hline\hspace{-10pt} 
\end{tabular} \vspace{-7pt} \\

\begin{tabular}{ccc}
\hspace{-10pt} 
\includegraphics[width=\imageWidth,height=\imageHeight]{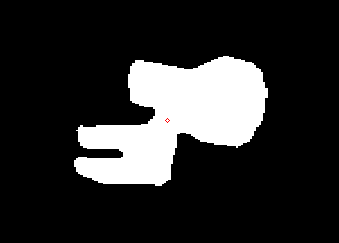} & \hspace{-15pt} 
\includegraphics[width=\imageWidth,height=\imageHeight]{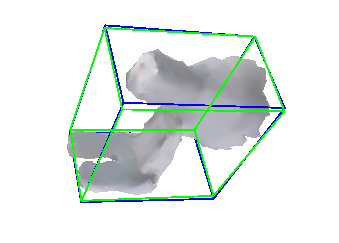} \hspace{-15pt} &
\includegraphics[width=\imageWidth,height=\imageHeight]{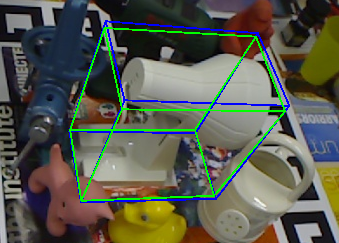} \hspace{-10pt} \\
\hline\hspace{-10pt} 
\end{tabular}

&
\begin{tabular}{ccc}
\hspace{-10pt} 
\includegraphics[width=\imageWidth,height=\imageHeight]{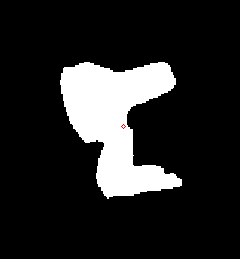} & \hspace{-15pt} 
\includegraphics[width=\imageWidth,height=\imageHeight]{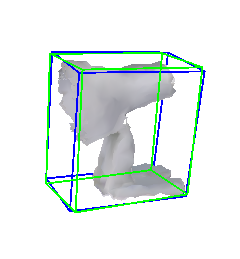} \hspace{-15pt} &
\includegraphics[width=\imageWidth,height=\imageHeight]{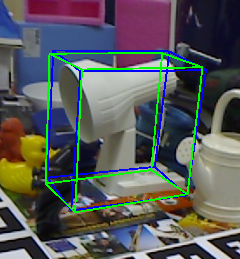} \hspace{-10pt} \\
\hline\hspace{-10pt} 
\end{tabular}

&
\begin{tabular}{ccc}
\hspace{-10pt} 
\includegraphics[width=\imageWidth,height=\imageHeight]{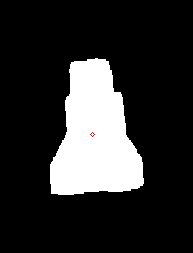} & \hspace{-15pt} 
\includegraphics[width=\imageWidth,height=\imageHeight]{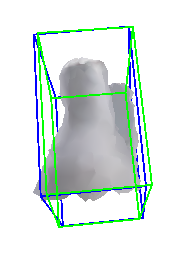} \hspace{-15pt} &
\includegraphics[width=\imageWidth,height=\imageHeight]{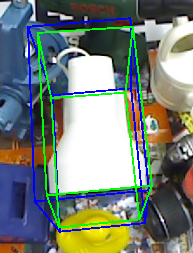} \hspace{-10pt} \\
\hline\hspace{-10pt} 
\end{tabular} \vspace{-7pt} \\

\begin{tabular}{ccc}
\hspace{-10pt} 
\includegraphics[width=\imageWidth,height=\imageHeight]{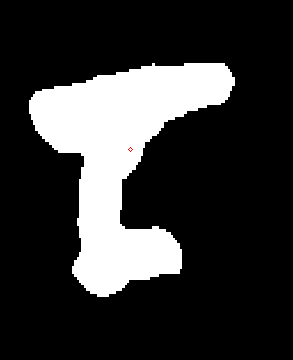} & \hspace{-15pt} 
\includegraphics[width=\imageWidth,height=\imageHeight]{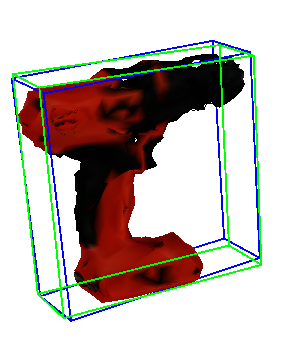} \hspace{-15pt} &
\includegraphics[width=\imageWidth,height=\imageHeight]{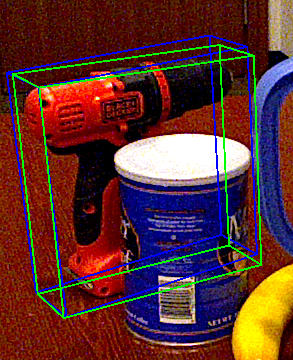} \hspace{-10pt} \\
\hline\hspace{-10pt} 
\end{tabular}

&
\begin{tabular}{ccc}
\hspace{-10pt} 
\includegraphics[width=\imageWidth,height=\imageHeight]{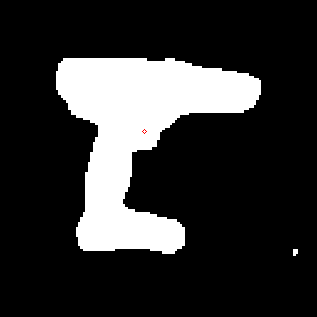} & \hspace{-15pt} 
\includegraphics[width=\imageWidth,height=\imageHeight]{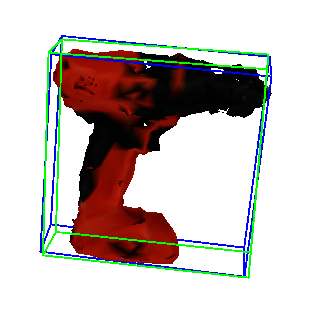} \hspace{-15pt} &
\includegraphics[width=\imageWidth,height=\imageHeight]{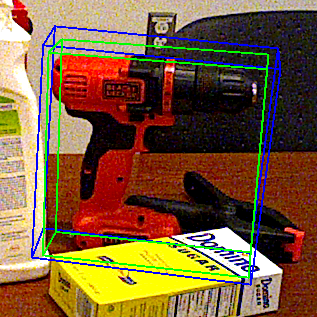} \hspace{-10pt} \\
\hline\hspace{-10pt} 
\end{tabular}

&
\begin{tabular}{ccc}
\hspace{-10pt} 
\includegraphics[width=\imageWidth,height=\imageHeight]{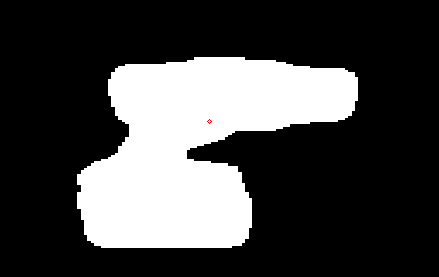} & \hspace{-15pt} 
\includegraphics[width=\imageWidth,height=\imageHeight]{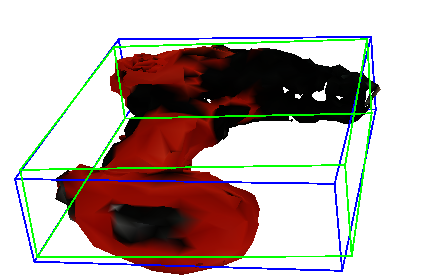} \hspace{-15pt} &
\includegraphics[width=\imageWidth,height=\imageHeight]{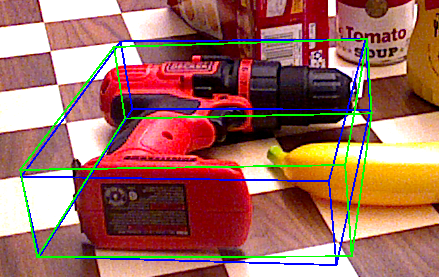} \hspace{-10pt} \\
\hline\hspace{-10pt} 
\end{tabular} \vspace{-7pt} \\

\begin{tabular}{ccc}
\hspace{-10pt} 
\includegraphics[width=\imageWidth,height=\imageHeight]{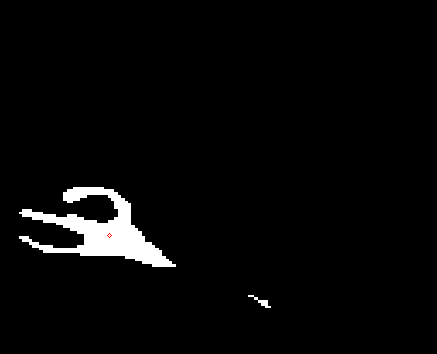} & \hspace{-15pt} 
\includegraphics[width=\imageWidth,height=\imageHeight]{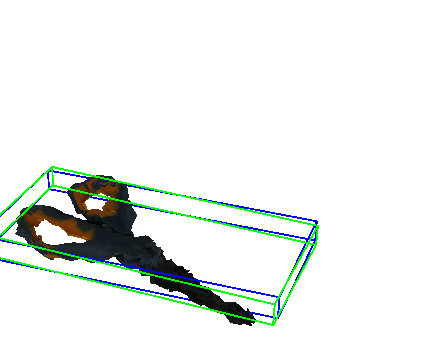} \hspace{-15pt} &
\includegraphics[width=\imageWidth,height=\imageHeight]{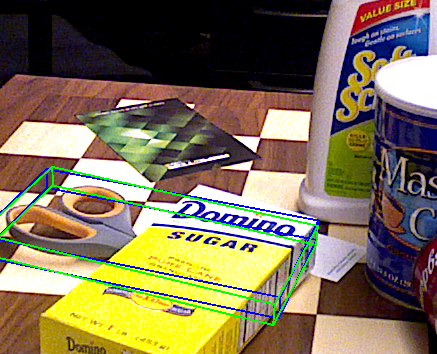} \hspace{-10pt} \\
\hspace{-8pt}(a) & (b) & \hspace{8pt}(c) \\
\end{tabular}

&
\begin{tabular}{ccc}
\hspace{-10pt} 
\includegraphics[width=\imageWidth,height=\imageHeight]{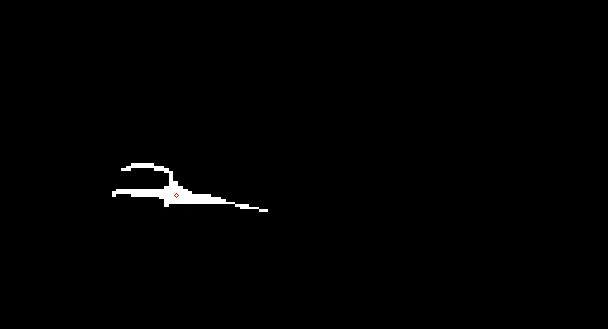} & \hspace{-15pt} 
\includegraphics[width=\imageWidth,height=\imageHeight]{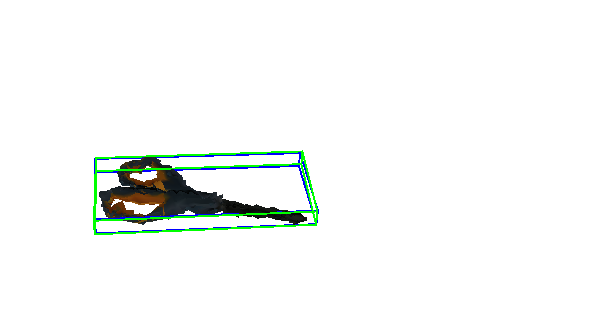} \hspace{-15pt} &
\includegraphics[width=\imageWidth,height=\imageHeight]{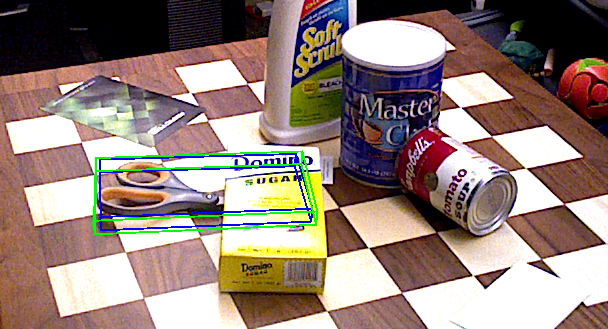} \hspace{-10pt} \\
\hspace{-8pt}(a) & (b) & \hspace{8pt}(c) \\
\end{tabular}

&
\begin{tabular}{ccc}
\hspace{-10pt} 
\includegraphics[width=\imageWidth,height=\imageHeight]{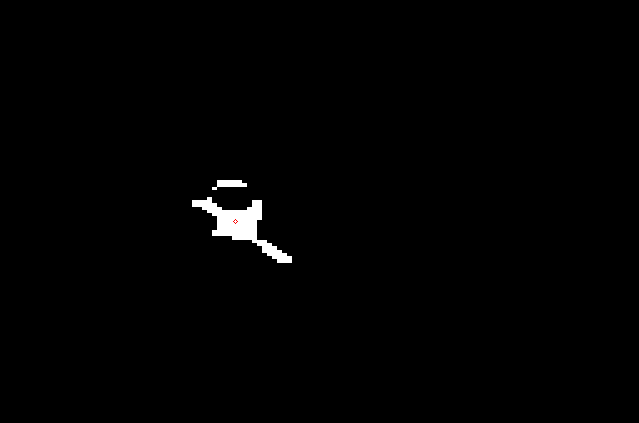} & \hspace{-15pt} 
\includegraphics[width=\imageWidth,height=\imageHeight]{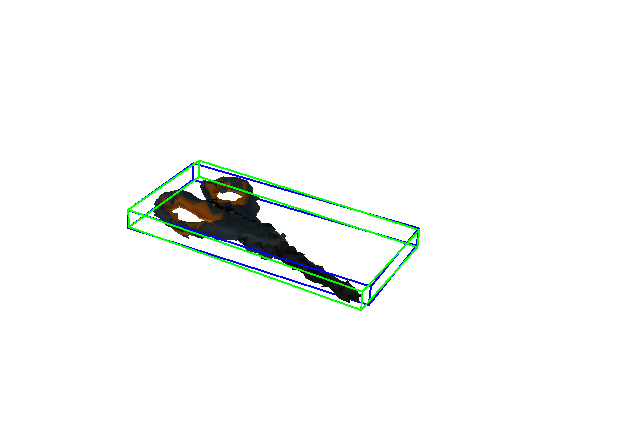} \hspace{-15pt} &
\includegraphics[width=\imageWidth,height=\imageHeight]{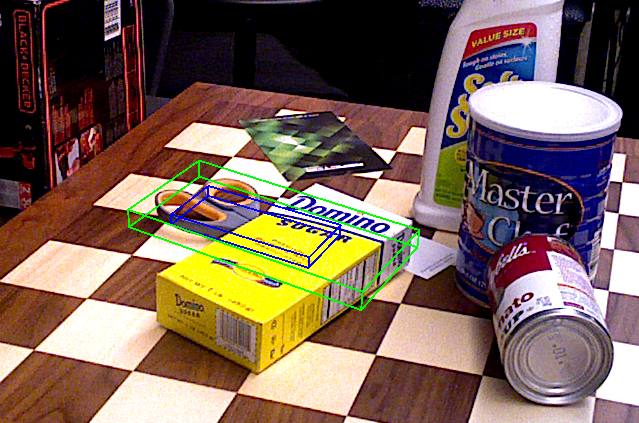} \hspace{-10pt} \\
\hspace{-8pt}(a) & (b) & \hspace{8pt}(c) \\
\end{tabular} \\
\end{tabular}
\caption{Qualitative results obtained with our method. Each row shows our results on examples of an object 
from the LINEMOD dataset with the last two rows showing examples from the YCB-Video dataset. 
For each test image, we provide (a) estimated mask and centroid projection, (b) reconstructed pose conditioned mesh and (c) groundtruth (green) and our estimated (blue) pose visualized on the input image. On the last column, we present failure examples which are the result of poor $\mask$ and $\centroid$ estimations or reconstruction errors due to occlusion, extreme illumination conditions or ambiguous viewpoints. Please note that the mask and the centroid projection are blended on the same image and test images are zoomed for better visualization. }
\label{fig:qualitative}
\end{figure*}

\subsection{3D Mesh Reconstruction}\label{reconstruction}

In the next stage of our pipeline, we reconstruct a pose conditioned 3D mesh of the target object. We employ a GraphCNN \cite{graphcnn}, which has been successfully applied to structure data modelling \cite{pix2mesh, handgraphcnn}, to output the 3D coordinates position of each mesh vertex. Assuming that for each rigid object, the topology of the object surface remains fixed, we can represent its surface as a pose conditioned 3D mesh $\meshcon$, and consequently, $\meshcon$ can be represented as an undirected graph $G=(V,E,W)$, where $V$ is a set of $|V(G)| = v$ vertices, $E$ a set of $|E(G)| = e$ undirected edges and $W = \{0,1\}^{v \times{} v}$ is an adjacency matrix, where $\{w_{ij}=1|(i,j) \in E, w_{ij} \in W\}$. In turn, our canonical mesh, i.e. without any applied transformation, can be represented as $G_{can}=(V_{can},E,W)$.

The outputs from the previous steps are encoded into a latent vector and then resized into a mesh signal $f \in \mathbb{R}^{|V| \times D}$, with $D$ representing the data dimensionality in each vertex. The signal is forward propagated through a hierarchical GraphCNN architecture \cite{graphcnn, handgraphcnn, facegraphcnn}. In its hierarchical structure graph convolution operations are intercalated with mesh feature upscaling, allowing us to operate on a finer mesh version. The output of the GraphCNN is the 3D position of each vertex. The full reconstructed mesh can be represented by $\hat{G}=(\hat{V}, E, W)$.
The pose conditioned 3D mesh reconstruction task is fully supervised by $L_G =  L_v + L_e +  L_l$, which combines multiple 3D loss constraints in the mesh generation process, similarly to \cite{handgraphcnn, pix2mesh}. The vertex loss $L_v$ constraints 3D vertex location, the edge loss $L_e$ penalizes incorrect edge length and the Laplacian loss $L_l$ enforces mesh surface smoothness:
\begin{equation}
L_v = \sum_{i=1}^{v} {\lVert { \hat{V}_i - V_i } \rVert}_{2}^{2}
\end{equation} 
\begin{equation}
L_e=\sum_{i}^{v}\sum_{V_k \in \mathcal{N}(V_i)}^{} \left( {\lVert{\hat{V}_i - \hat{V}_k} \rVert}_2^2 - {\lVert V_i - V_k \rVert}_2^2 \right) ^2
\end{equation}
\begin{equation}
L_l = \sum_{i=1}^{v} \left\lVert  (\hat{V}_i - V_i) - \frac{\sum_{V_k \in \mathcal{N}(V_i)} {(\hat{V}_k - V_k)}}{|\mathcal{N}(V_i)|} \right\rVert_2^2
\end{equation}
where $\mathcal{N}(V_i)$ indicates is the set of neighbouring vertices defined as $\{V_k \in \mathcal{N}(V_i) | W_{i,k} = 1\}$. 
In practice, we set the loss weighting parameters to $\lambda_v = 1.0$, $\lambda_e = 1.0$ and $\lambda_l = 1.0$.

\subsection{Allocentric Pose Estimation \& 6D Egocentric Pose Lifting}\label{lifting}

Given both meshes, we want to recover the allocentric orientation of the object by computing the optimal alignment rotation between the two sets of vertices. This task is known as the Generalized Orthogonal Proscrutes Problem \cite{procrustes} and in this context we formally define it as $\hat{V}(\hat{G}) = R_{a} V_{can}(G_{can})$, with $R_{a}$ relative rotation between the two set of vertices. The optimal solution for $R_{a}$ can be found by computing: 
\begin{equation}
    R_{a} = \mathcal{V} diag(1, 1,\dots, sign(det(\mathcal{U}\mathcal{V}^T)) \mathcal{U}^T
\end{equation}
where $\mathcal{U}, \mathcal{E}, \mathcal{V}^T = SVD\left(\hat{V}V_{can}^T\right)$. SVD can be implemented in a differentiable manner, therefore, we add an additional constraint, similarly to  \cite{keypoints}, the form of the pose loss $L_p$:

\vspace{-10pt}
\begin{equation}
    L_a = arcsin \left( \frac{ {\lVert \hat{R}_{a} - R_{a} \rVert}_2}{2 \sqrt{2}} \right)
\end{equation}
which penalizes the difference between the ground truth and estimated orientation. All losses from the previous stages and this stage, $L_s$, $L_g$ and $L_a$, are combined into a final loss $L$ to jointly learn parameters of the differentiable three stages of our pipeline by optimizing L:

\vspace{-5pt}
\begin{equation}
L = \lambda_s L_s + \lambda_g L_g + \lambda_p L_p
\end{equation}

We pay special attention to symmetric objects since different labeled poses might have identical appearances. For objects with symmetric properties, we follow the method proposed by \cite{nocs} and pre-define a object specific axis of symmetry. When computing $L_v$ and $L_a$, we generate additional groundtruth annotations by rotating $V$ and $R_a$ around the pre-defined axis of rotation. We measure the error for $L_v$ and $L_a$ on all available annotations and backpropagate through the minimum error.

At last, we lift the egocentric 6D object pose w.r.t the camera given the information recovered from previous stages. We start by positioning the estimated 2D centroid projection w.r.t. the detector's bounding box location. Following the approach described in \cite{3drcnn}, we compute $\pose = [\begin{smallmatrix} R&T\\ 0^T&1\end{smallmatrix}]$, where $R = R_c R_{a}$ and $T = R_c [0,0,\hat{d}]^T$. $R_c \in SO(3)$ is the rotation necessary to align the camera principal axis $a=[0,0,1]^T$ with $r=K_c^{-1}\hat{c}$ where $r$ is the ray passing through the repositioned 2D centroid projection and calculated with the inverse camera intrinsics matrix $K^{-1}$ and estimated 2D centroid projection $\hat{c}$. Then, $R_c$ can be calculated as $R_c=I + [v]_{\times} + \frac{[v]_{\times}^2}{1+a \cdot r}$ where $v = a \times r$ and $[v]_{\times}$ is the skew-symmetric cross-product matrix of $v$ and $I$ is the identity matrix.
In order to compute $T$, we need the distance to the object to the virtual ROI camera. Given the allocentric pose and the target object's 3D model, we render the object at an arbitrary distance $d$ with the estimated allocentric pose. Using the rendering and estimated masks' bounding box diagonal length, $l$ and $\hat{l}$ respectively, we can infer $\hat{d} = \frac{l \cdot d}{\hat{l}}$ \cite{ssd6d}.

\section{Experiments}

We evaluate our method and compare it against previous RGB based state of the art approaches\cite{seamless, bb8,ssd6d,posecnn,heatmaps}  that do not rely on iterative refinement stages or use of depth information for accurate 6D pose estimation. 

\subsection{Datasets}

\textbf{LINEMOD.} Since it was introduced, the LINEMOD dataset \cite{linemod} has been the most prolific dataset for 6D object pose estimation and has been the target of extensive benchmarking \cite{bb8, seamless, ssd6d, aae, posecnn, uncertainty}. The dataset is divided into 13 subsets specific for each object. 
Every image is annotated with its target object's rotation, translation and bounding box groundtruth, along with the camera intrinsic parameters. The authors also made available a full detailed 3D model of each object.

\textbf{OCCLUSION.} Additionally, we also evaluate our work on the more challenging OCCLUSION dataset. This dataset's images are from the \textit{Benchwise} LINEMOD scene but with all 9 other objects annotated and severely occluded. 

\textbf{YCB-Video.} The YCB-Video dataset was presented by Xiang \textit{et. al.} \cite{posecnn} in a response to the lack of real training data on the LINEMOD dataset. This dataset makes available 133,827 real and labeled images, which surpasses the available data on LINEMOD by two orders of magnitude. 

\subsection{Evaluation Metrics}

We evaluate and compare our approach by measuring ADD/ADI, 2D projection and AUC metrics on the three main benchmarking datasets for 6D object pose estimation LINEMOD\cite{linemod}, OCCLUSION\cite{occlusion} and YCB-Video\cite{posecnn}. 

\textbf{2D Projection. } The 2D Projection metric measures how close the 2D projected vertices are to the groundtruth, in pixel domain. We consider a correct pose if the mean 2D projection error is below $5px$. 

\textbf{ADD/ADI.} The ADD metric measures the mean Euclidean distance between the correspondent vertices of the estimated pose transformed model and the groundtruth pose transformed model. We consider a correct pose if the ADD score is below 10\% the target object's diameter. However, ADD heavily penalizes symmetric objects whose appearance is rotational invariant along the symmetry axis. For this reason, we measure ADI on objects which display symmetry properties, which will be denoted with *. Formally, these metrics are defined as:
\begin{equation}
    ADD = \frac{1}{|V|}\sum_{i=0}^{|V|} {\lVert (RV_i + T) - (\hat{R}V_i + \hat{T}) \rVert}_2
\end{equation}
\begin{equation}
    ADI = \frac{1}{|V|}\sum_{i=0}^{|V|} \min_{V_k \in V}{\lVert (RV_i + T) - (\hat{R}V_k + \hat{T}) \rVert}_2
\end{equation}

\textbf{AUC.} The AUC metric was first defined by Xiang \textit{et. al.} \cite{posecnn}. It is defined by the area under the accuracy-threshold curve when using the ADD/ADI metric. The curve is built by varying the threshold, to a maximum threshold of $10cm$. This metric is commonly not used on the LINEMOD and OCCLUSION datasets. For this reason, we only present AUC on the YCB-Video dataset.

From this point onward, we will refer to 2D Projection accuracy and ADD/ADI accuracy when discussing the percentage of correct poses using the previous metrics.

\begin{table}[t]
\setlength{\tabcolsep}{2pt}
\footnotesize
\centering
\begin{tabular}{|l|ccc|ccc|}
\hline
\multicolumn{1}{|c|}{Method}  & \multicolumn{3}{c|}{ADD/ADI accuracy}                                                                        & \multicolumn{3}{c|}{2D Projection accuracy}                                                                \\ \hline
\multicolumn{1}{|c|}{Objects} & BB8\cite{bb8}  & Tekin\cite{seamless} & \textbf{OURS}  & BB8\cite{bb8} & Tekin\cite{seamless} & \textbf{OURS} \\ \hline \hline
Ape                           & 27.90                           & 21.62                                 & \textbf{35.11} & 95.30                          & 92.10                                 & \textbf{98.10}                 \\
Benchvise                     & 62.00                           & 81.80                                 & \textbf{85.01} & 80.00                          & \textbf{95.06}                        & 88.20                          \\
Camera                        & 40.10                           & 36.57                                 & \textbf{45.46} & 80.9                           & \textbf{93.24}                        & 80.80                          \\
Can                           & 48.10                           & \textbf{68.80}       & 68.31                           & 84.10                          & \textbf{97.44}                        & 94.10                          \\
Cat                           & \textbf{45.20} & 41.82                                 & 41.98                           & 97.10                          & \textbf{97.41}                        & 97.10                          \\
Driller                       & 58.60                           & 63.51                                 & \textbf{72.73} & 74.10                          & 79.41                                 & \textbf{83.20}                 \\
Duck                          & 32.80                           & 27.23                                 & \textbf{35.57 }& 81.20                          & 94.65                                 & \textbf{96.70}                 \\
Eggbox*                       & 40.00                           & 69.58                                 & \textbf{75.80} & 87.90                          & \textbf{90.33}                        & 81.10                          \\
Glue*                         & 27.00                           & \textbf{80.02}       & 53.40                           & 89.0                           & \textbf{96.53}                        & 67.60                          \\
Holepuncher                   & 42.40                           & 42.63                                 & \textbf{44.46} & 90.50                          & 92.86                                 & \textbf{97.30}                 \\
Iron                          & 67.00                           & 74.97                                 & \textbf{81.25} & 78.90                          & \textbf{82.94}                        & 82.50                          \\
Lamp                          & 39.90                           & 71.11                                 & \textbf{75.71} & 74.40                          & 76.87                                 & \textbf{82.60}                 \\
Phone                         & 35.20                           & 47.74                                 & \textbf{55.35} & 77.60                          & \textbf{86.07}                        & 83.70                          \\ \hline \hline
Average                       & 43.55                           & 55.95                                 & \textbf{59.32} & 83.9                           & \textbf{90.37}                        & 87.15                          \\ \hline
\end{tabular}
\vspace{5pt}
\caption{ADD/ADI comparison of our approach on the LINEMOD dataset with state-of-the-art. }
\label{table:eval}
\vspace{-10pt}
\end{table}

\subsection{Implementation Details}

In order to train our model on the LINEMOD and OCCLUSION datasets, we generate synthetic data from existing real images of the LINEMOD~\cite{linemod} dataset. Following the literature \cite{seamless, bb8}, we sample 15\% of each object's images as the training set and the remaining images are used for evaluation purposes. The image sampling is done so that the selected images have a relative pose between each other larger than $\frac{\pi}{12}$ \cite{bb8}. 
We crop the objects according to their silhouettes and place them on random backgrounds, utilizing the "Cut and Paste" technique \cite{cutpaste}. 
The YCB-Video dataset provides its own training set \cite{posecnn}.
Online augmentation is also performed, on all datasets, by randomly tweaking the hue, saturation and exposure of the image as well as the addition of Gaussian Noise.

We set the loss weighting parameters to $\lambda_s = 1.0$, $\lambda_g = 1.0$ and $\lambda_l = 5.0$. Our networks' parameters are updated for 20 epochs using the Adam \cite{adam} optimizer, with its initial learning rate set to $10^{-3}$ and batch-size of $8$. 

Given an input image $480 \times 640$, our method runs at 17 FPS on a GTX1080 Ti GPU. Our detection 
is the most costly operation, taking around 28ms to compute. Mask segmentation and centroid estimation take 8ms while the GraphCNN forward pass takes 14ms. Finally, the lifting to 6D takes 9ms, totalling at 59ms for the whole computation.

\bgroup
\def\arraystretch{1.45}
\begin{table}[t]
\begin{center}
    \footnotesize
    \setlength{\tabcolsep}{4pt}
    \begin{tabular}{|l|c|c|c|c|}
    \hline
    \multicolumn{1}{|c|}{Methods} & PoseCNN\cite{posecnn} & Tekin\cite{seamless} &  \textbf{Ours} & Oberweger\cite{heatmaps}    \\ \hline \hline
    Ape                          & 34.6                         & 7.01                       & 42.50                              & \textbf{64.7} \\
    Benchwise$^\dagger$                          & -                         & -                       & \textbf{87.73}                              & - \\
    Can                          & 15.1                         & 11.2                       & 30.72                              & \textbf{53.0} \\
    Cat                          & 10.4                         & 3.62                       & 22.32                              & \textbf{47.9} \\
    Driller                      & 7.40                         & 1.40                       & \textbf{44.89}                     & 35.1          \\
    Duck                         & 31.8                         & 5.07                       & 16.80                              & \textbf{36.1} \\
    Eggbox*                      & 1.90                         & -                          & \textbf{15.57}                     & 10.3          \\
    Glue*                        & 13.8                         & 6.53                       & 4.28                               & \textbf{44.9} \\
    Holepuncher                  & 23.1                         & 8.26                       & \textbf{57.5}                      & 52.9          \\ \hline \hline
    Average w/o $\dagger$                     & \multicolumn{1}{c}{17.2}    & \multicolumn{1}{|c}{6.16}  & \multicolumn{1}{|c|}{29.32}         & \textbf{43.1} \\ \hline
    \end{tabular}
    \centering
    \vspace{5pt}
    \caption{2D Projection accuracy results on the OCCLUSION dataset.}
    \label{table:occlusion}
\end{center}
\vspace{-10pt}
\vspace{-12pt}
\end{table}
\egroup

\subsection{Comparison to State of the Art}

\begin{table}[b]
\vspace{-5pt}
    \centering
    \footnotesize
    \setlength{\tabcolsep}{3pt}
    \begin{tabular}{|c|ccc|}
    \hline
    Methods       & \multicolumn{1}{c}{PoseCNN\cite{posecnn}} & \multicolumn{1}{c}{Oberweger\cite{heatmaps}} & \textbf{OURS} \\ \hline \hline
    ADD/ADI       & 21.30                         & 33.60                           & \textbf{47.09}     \\ \hline
    AUC           & 61.00                         & 61.40                           & \textbf{67.52}          \\ \hline
    2D Projection & 3.72                         & 23.10                           & \textbf{55.22}         \\ \hline
    \end{tabular}
    \vspace{5pt}
    \caption{Comparison of YCB-Video dataset results with state-of-the-art approaches.}
    \label{table:ycb}
\end{table}
    
 \textbf{LINEMOD.} We report higher ADD/ADI accuracies for all but 3 objects and achieve an overall average accuracy $\sim 4\%$ higher than Tekin\cite{seamless}, as can be observed in Table \ref{table:eval}. However, we fall short of having the same results when evaluating our method using the 2D Projection accuracy. This poorer performance is due to the lack of precision by our 2D localisation tasks. While our pose estimation is done in 3D space, YOLO6D \cite{seamless} perform keypoint localization in 2D space. For this reason, we outperform in metrics that reward 3D precision. 
We present additional qualitative results on Figure \ref{fig:qualitative}. 

\textbf{OCCLUSION.} The results on the OCCLUSION dataset are presented on Table \ref{table:occlusion}, with the results from BB8 and Tekin being reported in \cite{heatmaps}. The presented results show competitive scores when compared with prior works. We outperform BB8 and Tekin by 12.2\% and 23.16\% respectively. However, the occlusion aware approach of \cite{heatmaps} is able to surpass our method. 
Our object topology based pose estimation approach outperforms the more recent work of Wang \textit{et. al.} \cite{nocs}, even though this method requires RGB-D input, by 5.61\% when additionally considering the \textit{Benchwise} object, having an overall 35.81\% 2D Projection accuracy. 

\textbf{YCB-Video.} Additionally, we present the results of our method on the more recent YCB-Video dataset. Our results significantly surpass both PoseCNN~\cite{posecnn} and Oberweger~\cite{heatmaps} on all accounted metrics. Due to not being designed for occluded scenarios only, our method can better generalize for many different scenes by leveraging it's 3D reasoning with precise localization, which leads to a more accurate pose estimation estimation. A detail look at the our method's performance on this dataset can be seen on Table \ref{table:ycb} with qualitative examples being shown on Figure \ref{fig:qualitative}.

\begin{figure*}[t]
    \begin{minipage}[b]{.39\linewidth}
        \centering
        \footnotesize
        \setlength{\tabcolsep}{1pt}
        \begin{tabular}{|l|cc|cc|c|}
        \hline
        \multicolumn{1}{|c|}{Information}  & \multicolumn{2}{c|}{w/o Priors}                                                                        & \multicolumn{3}{c|}{Ours w/ Priors}   \\
        \hline
        \multicolumn{1}{|c|}{Objects}                 & \textbf{Ours} & SSD6D\cite{ssd6d}  & w/ $\centroidgt$  & w/ $\maskgt$ & w/ $\centroidgt$\&$\maskgt$  \\ 
        \hline \hline
        Ape                                                 & 35.11& 18.04 & 35.52                    & 51.13                    & 53.16                      \\
        Benchwise                                           & 85.01& 67.38 & 85.83                     & 86.49                      & 87.07                     \\
        Camera                                              & 45.46& 27.81 & 48.38                     & 70.52                      & 73.61                      \\
        Can                                                 & 68.31& 64.30 & 69.57                     & 80.18                      & 81.35                      \\
        Cat                                                 & 41.98& 33.33 & 42.49                     & 58.44                      & 59.46                     \\
        Driller                                             & 72.73& 62.04 & 72.90                     & 75.42                     & 76.60                      \\
        Duck                                                & 35.57& 16.67 & 36.13                     & 55.98                      & 57.26                      \\
        Eggbox*                                             & 75.8 & 75.20 & 91.90                     & 97.50                      & 97.60                      \\
        Glue*                                               & 53.4 & 52.50 & 56.00                     & 56.10                      & 56.60                      \\
        Holepuncher                                         & 44.46 & 35.57 & 45.43                     & 62.57                      & 63.62                     \\
        Iron                                                &  81.25 & 62.59 & 81.77                     & 79.08                      & 79.51                      \\
        Lamp                                                & 75.71 & 70.09 & 76.12                     & 83.70                      & 84.60                      \\
        Phone                                               & 55.35 & 43.36 & 56.72                     & 69.75                      & 70.64                      \\ \hline \hline 
        Average                        & \textbf{59.32} & 48.37 & 61.44 & \textbf{71.30} & \textbf{72.39} \\ \hline
        \end{tabular}
        \vspace{5pt}
        \caption{ADD/ADI comparison of our approach by using groundtruth priors for the centroid and the mask estimation intermediate steps on LINEMOD.}
        \label{table:limits}
\vspace{-5pt}
    \end{minipage}
    \qquad
    \begin{minipage}[t]{\linewidth}
    \begin{minipage}[b]{.18\linewidth}
        \centering
        
        \includegraphics[width=\linewidth, height=5.4cm]{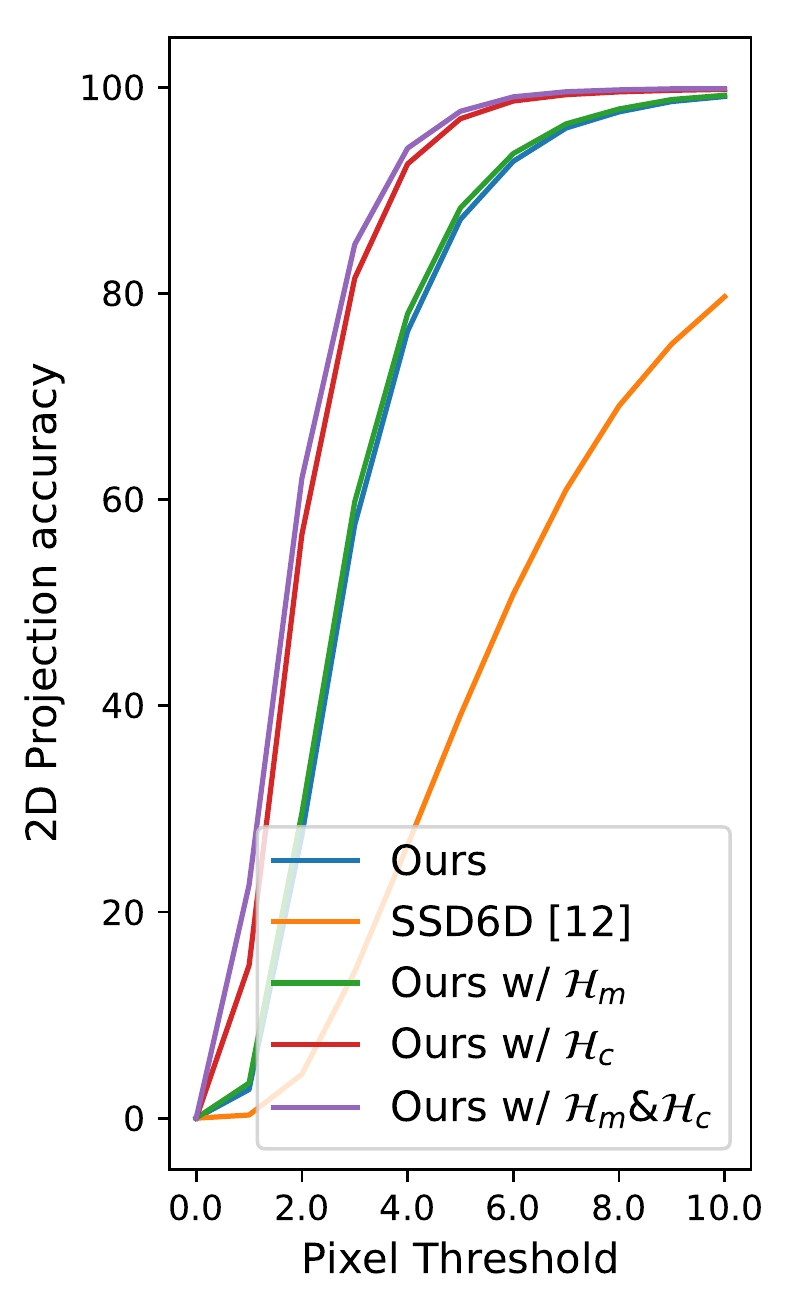}
        \caption{2D Projection accuracy using different priors.}
    \label{fig:long}
\vspace{-5pt}
    \end{minipage}
    \qquad
    \begin{minipage}[b]{.35\linewidth}
        \centering
        \includegraphics[width=\linewidth]{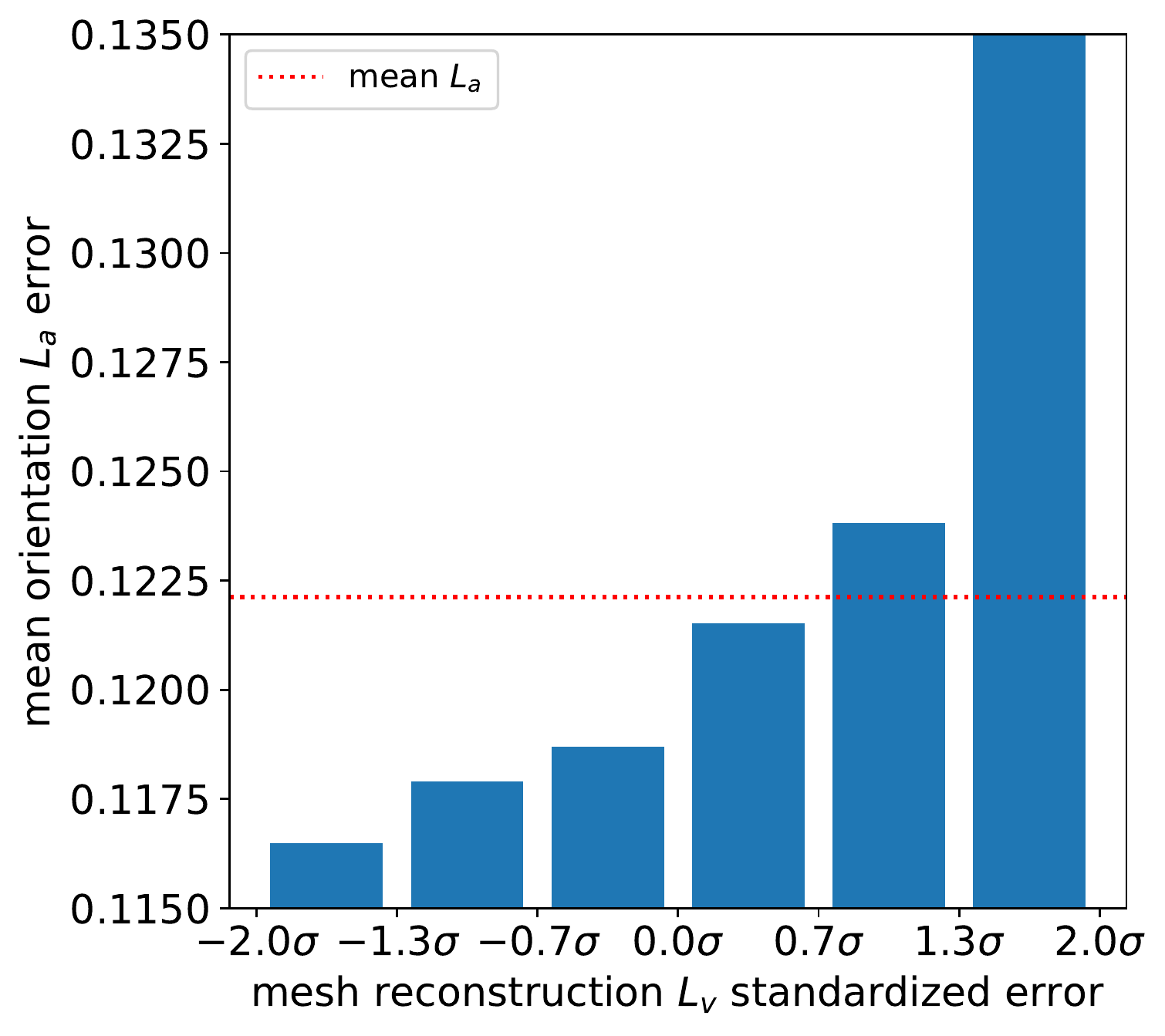}
        \caption{Histogram of mean pose losses $L_a$ over standardized mesh reconstruction error $L_v$ thresholds.}
        \label{fig:histogram}
\vspace{-5pt}
    \end{minipage}
    \end{minipage}
    
\end{figure*}

\section{Discussions} \label{sec:discussions}

In this section, we show and discuss the individual impact of each stage of our pipeline in the final 6D pose estimation. Experiments on this section were made with the LINEMOD dataset.

\subsection{2D Localization} 
As previously discussed, the use of a detector and both $\mask$ and $\centroid$ estimations are \textit{ad hoc} components in our full 6D pose estimator. In Table \ref{table:limits}, we present our model's ADD/ADI accuracy when groundtruth information is available for $\mask$ and $\centroid$ and on Figure \ref{fig:long} each respective 2D Projection accuracy. In \cite{ssd6d}, the authors make use of their orientation and bounding box estimations in order to recover the centroid position. We show that our method of estimating the 2D centroid projection outperforms the one used by \cite{ssd6d} on all objects. We are able to more precisely locate the centroid's projection due to our high resolution $\centroid$ estimation.
When using non-available groundtruth information on secondary tasks, our method achieves extremely high accuracy on most objects. We can also see the impact that each 2D location task has on each of the metrics. 2D Projection accuracy is highly influenced by the quality of $\centroid$ because it determines where the object will be projected in the 2D plane of the image. On the other hand, ADD/ADI is correlated with the quality of the $\mask$  since it determines the distance of the object to the camera, therefore more heavily impacting $\pose$ in 3D space.

\subsection{Mesh Reconstruction} 
In addition to measuring the quality of the 2D detection tasks, we also evaluate more precisely the mesh reconstruction errors. In order to correctly recover the allocentric pose, the reconstructed mesh should match its canonical counterpart. Otherwise, the alignment strategy will be unable to recover the correct orientation. Therefore, we want to measure how the quality of the reconstructed mesh affects the orientation recovery. In an effort to quantify this relation, we measure the mean orientation error $L_a$ for meshes with a reconstruction error $L_v$ between certain thresholds. Since $L_v$ fluctuates among objects, we standardize it for each object before studying the correlation. 

The histogram in Figure \ref{fig:histogram} measures the impact of the reconstruction quality in the pose prediction. When the reconstruction error is either in the top or lower $\sim 16$ percentile, we observe a correspondent high or low pose error. We can conclude that the quality of the recovered orientation is correlated with the quality of the reconstruction mesh. Taking this observation into account allows our model to have an intrinsic confidence score on the recovered pose simply by measuring the reconstruction error. Since the reconstruction error can be computed without any groundtruth information, our model is capable of self validating its orientation estimation at test time.

\section{Conclusions and Future Work}
In this work, we introduced a new method for instance-level 6D object pose, which takes full advantage of the target object's topological information. Our method extracts both the object's mask and 2D centroid projection with high resolution to precisely detect the object in 2D. We perform estimation in 3D space by reconstructing a 3D pose conditioned mesh of the target from which we recover its orientation by aligning it with the canonical version of the target's 3D mesh. Our method outperforms state-of-the-art on the standard 6D object pose dataset LINEMOD. Additionally, we present our method's results on more challenging datasets, OCCLUSION and YCB-Video, outperforming previous non-occlusion aware approaches while being comparable to occlusion aware methods.
We also show we can use the reconstruction error as an indicator of pose estimation precision which allows our model to self validate at test time, which is crucial on real life applications.\\

\noindent {\large \textbf{Acknowledgements.}} This work was in part financially supported via student consultation for Remark Holdings, Inc.

{\small
\bibliographystyle{ieee}
\bibliography{strings,egbib}

\begin{thebibliography}{10}\itemsep=-1pt

\bibitem{densebody}
R.~Alp~G{\"u}ler, N.~Neverova, and I.~Kokkinos.
\newblock {DensePose}: Dense human pose estimation in the wild.
\newblock In {\em Conference on Computer Vision and Pattern Recognition}, 2018.

\bibitem{facedense2}
R.~Alp~Guler, G.~Trigeorgis, E.~Antonakos, P.~Snape, S.~Zafeiriou, and
  I.~Kokkinos.
\newblock Densereg: Fully convolutional dense shape regression in-the-wild.
\newblock In {\em The IEEE Conference on Computer Vision and Pattern
  Recognition (CVPR)}, July 2017.

\bibitem{sb}
S.~Baek, K.~I. Kim, and T.-K. Kim.
\newblock Pushing the envelope for rgb-based dense 3d hand pose estimation via
  neural rendering.
\newblock In {\em Proceedings of the IEEE Conference on Computer Vision and
  Pattern Recognition (CVPR)}, pages 1067--1076, 2019.

\bibitem{tripletpose}
V.~Balntas, A.~Doumanoglou, C.~Sahin, J.~Sock, R.~Kouskouridas, and T.-K. Kim.
\newblock Pose guided {RGBD} feature learning for {3D} object pose estimation.
\newblock In {\em International Conference on Computer Vision}, 2017.

\bibitem{keypoints1}
E.~Brachmann, A.~Krull, F.~Michel, S.~Gumhold, J.~Shotton, and C.~Rother.
\newblock Learning {6D} object pose estimation using {3D} object coordinates.
\newblock In {\em European Conference on Computer Vision}, 2014.

\bibitem{occlusion}
E.~Brachmann, A.~Krull, F.~Michel, S.~Gumhold, J.~Shotton, and C.~Rother.
\newblock Learning 6d object pose estimation using 3d object coordinates.
\newblock In {\em Proceedings of the European Conference on Computer Vision
  (ECCV)}, 2014.

\bibitem{uncertainty}
E.~Brachmann, F.~Michel, A.~Krull, M.~Ying~Yang, S.~Gumhold, et~al.
\newblock Uncertainty-driven 6d pose estimation of objects and scenes from a
  single rgb image.
\newblock In {\em Proceedings of the IEEE Conference on Computer Vision and
  Pattern Recognition (CVPR)}, 2016.

\bibitem{depth1}
C.~Choi and H.~I. Christensen.
\newblock {RGB-D} object pose estimation in unstructured environments.
\newblock {\em Robotics and Autonomous Systems}, 75:595--613, 2016.

\bibitem{graphcnn}
M.~Defferrard, X.~Bresson, and P.~Vandergheynst.
\newblock Convolutional neural networks on graphs with fast localized spectral
  filtering.
\newblock In {\em Advances in Neural Information Processing Systems}, 2016.

\bibitem{andreas}
A.~Doumanoglou, R.~Kouskouridas, S.~Malassiotis, and T.-K. Kim.
\newblock Recovering 6d object pose and predicting next-best-view in the crowd.
\newblock In {\em The IEEE Conference on Computer Vision and Pattern
  Recognition (CVPR)}, June 2016.

\bibitem{cutpaste}
D.~Dwibedi, I.~Misra, and M.~Hebert.
\newblock {Cut, Paste and Learn}: Surprisingly easy synthesis for instance
  detection.
\newblock In {\em International Conference on Computer Vision}, 2017.

\bibitem{denseface}
Y.~Feng, F.~Wu, X.~Shao, Y.~Wang, and X.~Zhou.
\newblock Joint 3d face reconstruction and dense alignment with position map
  regression network.
\newblock In {\em Proceedings of the European Conference on Computer Vision
  (ECCV)}, 2018.

\bibitem{handgraphcnn}
L.~Ge, Z.~Ren, Y.~Li, Z.~Xue, Y.~Wang, J.~Cai, and J.~Yuan.
\newblock {3D} hand shape and pose estimation from a single {RGB} image.
\newblock {\em arXiv preprint arXiv:1903.00812}, 2019.

\bibitem{shapelearning}
P.~Henderson and V.~Ferrari.
\newblock Learning single-image {3D} reconstruction by generative modelling of
  shape, pose and shading.
\newblock {\em arXiv preprint arXiv:1901.06447}, 2019.

\bibitem{linemod}
S.~{Hinterstoisser}, S.~{Holzer}, C.~{Cagniart}, S.~{Ilic}, K.~{Konolige},
  N.~{Navab}, and V.~{Lepetit}.
\newblock Multimodal templates for real-time detection of texture-less objects
  in heavily cluttered scenes.
\newblock In {\em International Conference on Computer Vision}, 2011.

\bibitem{bop}
T.~Hodan, F.~Michel, E.~Brachmann, W.~Kehl, A.~GlentBuch, D.~Kraft, B.~Drost,
  J.~Vidal, S.~Ihrke, X.~Zabulis, et~al.
\newblock Bop: benchmark for 6d object pose estimation.
\newblock In {\em Proceedings of the European Conference on Computer Vision
  (ECCV)}, 2018.

\bibitem{segmentationdriven}
Y.~Hu, J.~Hugonot, P.~Fua, and M.~Salzmann.
\newblock Segmentation-driven {6D} object pose estimation.
\newblock {\em arXiv preprint arXiv:1812.02541}, 2018.

\bibitem{hausdorff}
D.~P. Huttenlocher, W.~J. Rucklidge, and G.~A. Klanderman.
\newblock Comparing images using the hausdorff distance under translation.
\newblock In {\em Conference on Computer Vision and Pattern Recognition}, 1992.

\bibitem{ssd6d}
W.~Kehl, F.~Manhardt, F.~Tombari, S.~Ilic, and N.~Navab.
\newblock {SSD-6D}: Making {RGB}-based {3D} detection and {6D} pose estimation
  great again.
\newblock In {\em International Conference on Computer Vision}, 2017.

\bibitem{posenet}
A.~Kendall, M.~Grimes, and R.~Cipolla.
\newblock Posenet: A convolutional network for real-time 6-dof camera
  relocalization.
\newblock In {\em The IEEE International Conference on Computer Vision (ICCV)},
  December 2015.

\bibitem{adam}
D.~P. Kingma and J.~Ba.
\newblock Adam: A method for stochastic optimization.
\newblock In {\em International Conference on Learning Representations (ICLR)},
  2015.

\bibitem{3drcnn}
A.~Kundu, Y.~Li, and J.~M. Rehg.
\newblock {3D-RCNN}: Instance-level {3D} object reconstruction via
  render-and-compare.
\newblock In {\em Conference on Computer Vision and Pattern Recognition}, 2018.

\bibitem{epnp}
V.~Lepetit, F.~Moreno-Noguer, and P.~Fua.
\newblock Epnp: An accurate o (n) solution to the pnp problem.
\newblock {\em International journal of computer vision}, 81(2):155, 2009.

\bibitem{chamfer}
M.-Y. Liu, O.~Tuzel, A.~Veeraraghavan, and R.~Chellappa.
\newblock Fast directional chamfer matching.
\newblock In {\em Conference on Computer Vision and Pattern Recognition}, 2010.

\bibitem{ssd}
W.~Liu, D.~Anguelov, D.~Erhan, C.~Szegedy, S.~Reed, C.-Y. Fu, and A.~C. Berg.
\newblock {SSD}: Single shot multibox detector.
\newblock In {\em European Conference on Computer Vision}, 2016.

\bibitem{multiscale}
S.~Nah, T.~Hyun~Kim, and K.~Mu~Lee.
\newblock Deep multi-scale convolutional neural network for dynamic scene
  deblurring.
\newblock In {\em Proceedings of the IEEE Conference on Computer Vision and
  Pattern Recognition (CVPR)}, 2017.

\bibitem{heatmaps}
M.~Oberweger, M.~Rad, and V.~Lepetit.
\newblock Making deep heatmaps robust to partial occlusions for {3D} object
  pose estimation.
\newblock In {\em European Conference on Computer Vision}, 2018.

\bibitem{bb8}
M.~Rad and V.~Lepetit.
\newblock {BB8}: a scalable, accurate, robust to partial occlusion method for
  predicting the {3D} poses of challenging objects without using depth.
\newblock In {\em International Conference on Computer Vision}, 2017.

\bibitem{facegraphcnn}
A.~Ranjan, T.~Bolkart, S.~Sanyal, and M.~J. Black.
\newblock Generating {3D} faces using convolutional mesh autoencoders.
\newblock In {\em European Conference on Computer Vision}, 2018.

\bibitem{yolo}
J.~Redmon, S.~Divvala, R.~Girshick, and A.~Farhadi.
\newblock {You Only Look Once}: Unified, real-time object detection.
\newblock In {\em Conference on Computer Vision and Pattern Recognition}, 2016.

\bibitem{fasterrcnn}
S.~Ren, K.~He, R.~Girshick, and J.~Sun.
\newblock Faster {R-CNN}: Towards real-time object detection with region
  proposal networks.
\newblock In {\em Advances in Neural Information Processing Systems}, 2015.

\bibitem{local_descriptors}
F.~Rothganger, S.~Lazebnik, C.~Schmid, and J.~Ponce.
\newblock {3D} object modeling and recognition using local affine-invariant
  image descriptors and multi-view spatial constraints.
\newblock {\em International Journal of Computer Vision}, 66(3):231--259, 2006.

\bibitem{procrustes}
P.~H. Sch{\"o}nemann.
\newblock A generalized solution of the orthogonal procrustes problem.
\newblock {\em Psychometrika}, 31(1):1--10, 1966.

\bibitem{depth_only}
J.~Sock, K.~I. Kim, C.~Sahin, and T.-K. Kim.
\newblock Multi-task deep networks for depth-based {6D} object pose and joint
  registration in crowd scenarios.
\newblock In {\em British Machine Vision Conference}, 2018.

\bibitem{aae}
M.~Sundermeyer, Z.-C. Marton, M.~Durner, M.~Brucker, and R.~Triebel.
\newblock Implicit {3D} orientation learning for {6D} object detection from
  {RGB} images.
\newblock In {\em European Conference on Computer Vision}, 2018.

\bibitem{keypoints}
S.~Suwajanakorn, N.~Snavely, J.~J. Tompson, and M.~Norouzi.
\newblock Discovery of latent {3D} keypoints via end-to-end geometric
  reasoning.
\newblock In {\em Advances in Neural Information Processing Systems}, 2018.

\bibitem{aly}
A.~Tejani, D.~Tang, R.~Kouskouridas, and T.-K. Kim.
\newblock Latent-class hough forests for 3d object detection and pose
  estimation.
\newblock In {\em Proceedings of the European Conference on Computer Vision
  (ECCV)}, 2014.

\bibitem{seamless}
B.~Tekin, S.~N. Sinha, and P.~Fua.
\newblock {Real-Time} seamless single shot {6D} object pose prediction.
\newblock In {\em Conference on Computer Vision and Pattern Recognition}, 2018.

\bibitem{densehand}
C.~Wan, T.~Probst, L.~Van~Gool, and A.~Yao.
\newblock Dense {3D} regression for hand pose estimation.
\newblock In {\em Conference on Computer Vision and Pattern Recognition}, 2018.

\bibitem{densefusion}
C.~Wang, D.~Xu, Y.~Zhu, R.~Mart{\'\i}n-Mart{\'\i}n, C.~Lu, L.~Fei-Fei, and
  S.~Savarese.
\newblock {DenseFusion}: {6D} object pose estimation by iterative dense fusion.
\newblock {\em arXiv preprint arXiv:1901.04780}, 2019.

\bibitem{nocs}
H.~Wang, S.~Sridhar, J.~Huang, J.~Valentin, S.~Song, and L.~J. Guibas.
\newblock Normalized object coordinate space for category-level {6D} object
  pose and size estimation.
\newblock {\em arXiv preprint arXiv:1901.02970}, 2019.

\bibitem{pix2mesh}
N.~Wang, Y.~Zhang, Z.~Li, Y.~Fu, W.~Liu, and Y.-G. Jiang.
\newblock {Pixel2Mesh}: Generating {3D} mesh models from single {RGB} images.
\newblock In {\em European Conference on Computer Vision}, 2018.

\bibitem{posecnn}
Y.~Xiang, T.~Schmidt, V.~Narayanan, and D.~Fox.
\newblock {PoseCNN}: A convolutional neural network for {6D} object pose
  estimation in cluttered scenes.
\newblock {\em Robotics: Science and Systems (RSS)}, 2018.

\end{thebibliography}
}

\end{document}